\documentclass[11pt,a4paper]{article}
\usepackage[hyperref]{acl2019}
\usepackage{times}
\usepackage{latexsym}
\usepackage{booktabs}
\usepackage{amsmath, amssymb}
\usepackage{url}
\usepackage{graphicx}
\usepackage{multirow}
\usepackage{makecell}
\usepackage{array}

\DeclareMathOperator\softmax{\operatorname{softmax}}

\makeatletter
\newcommand*\bigcdot{\mathpalette\bigcdot@{.5}}
\newcommand*\bigcdot@[2]{\mathbin{\vcenter{\hbox{\scalebox{#2}{$\m@th#1\bullet$}}}}}
\makeatother

\aclfinalcopy 
\def\aclpaperid{2701} 

\title{Generating  Summaries with \\Topic Templates and Structured Convolutional Decoders}

 \author{Laura Perez-Beltrachini \hspace{1cm} Yang Liu \hspace{1cm} Mirella Lapata \\
Institute for Language, Cognition and Computation \\
School of Informatics, University of Edinburgh \\
10 Crichton Street, Edinburgh EH8 9AB \\
   {\tt \{lperez,mlap\}@inf.ed.ac.uk}\hspace{.5cm} {\tt yang.liu2@ed.ac.uk}}

\date{}

\makeatletter
\newcommand{\thickhline}{%
    \noalign {\ifnum 0=`}\fi \hrule height 1pt
    \futurelet \reserved@a \@xhline
}
\makeatother

\begin{document}
\maketitle
\begin{abstract}
  Existing neural generation approaches create multi-sentence text as
  a single sequence. In this paper we propose a structured
  convolutional decoder that is guided by the content structure of
  target summaries.  We compare our model with existing sequential
  decoders on three data sets representing different domains.
  Automatic and human evaluation demonstrate that our summaries have
  better content coverage.
\end{abstract}

\section{Introduction}

Abstractive multi-document summarization aims at generating a coherent
summary from a cluster of thematically related documents.  Recently,
\citeauthor{liu2018generating} (\citeyear{liu2018generating}) proposed
generating the lead section of a Wikipedia article as a variant of
multi-document summarization and released WikiSum, a large-scale
summarization dataset which enables the training of neural
models. 

Like most previous work on neural text generation
\cite{gardent-EtAl:2017:INLG2017,see2017get,wiseman2017challenges,puduppully2018data,Celikyilmaz2018,liu2018generating,wikigen2018,gcn2text2018},
\citet{liu2018generating} represent the target summaries as a single
long sequence, despite the fact that documents are organized into
topically coherent text segments, exhibiting a specific structure in
terms of the content they discuss \cite{barzilay2004catching}. This is
especially the case when generating text within a specific domain
where certain topics might be discussed in a specific order
\cite{Wray:2002}. For instance, the summary in Table~\ref{tab:task} is
about a species of damselfly; the second sentence describes the
\emph{region} where the species is found and the fourth the \emph{type
  of habitat} the species lives in. We would expect other Animal
Wikipedia summaries to exhibit similar content organization.

In this work we propose a neural model which is guided by the topic
structure of target summaries, i.e.,~the way content is organized into
sentences and the type of content these sentences discuss.  Our model
consists of a structured decoder which is trained to predict a
sequence of sentence topics that should be discussed in the summary
and to generate sentences based on these.  We extend the convolutional
decoder of \citeauthor{gehring2017convs2s}
(\citeyear{gehring2017convs2s}) so as to be aware of which topics to
mention in each sentence as well as their position in the target
summary. We argue that a decoder which \emph{explicitly} takes
content structure into account could lead to better summaries and
alleviate well-known issues with neural generation models being too
general, too brief, or simply incorrect.

\begin{table*}[ht]
{\scriptsize
\begin{tabular}{p{15cm}}
 agriocnemis zerafica is a species of damselfly in the family coenagrionidae.   
 it is native to africa, where it is widespread across the central and western 
 nations of the continent. it is known by the common name sahel wisp.   
 this species occurs in swamps and pools in dry regions.   
 there are no major threats but it may be affected by pollution and habitat loss 
 to agriculture and development. \\
 \hline
 \hline
 {\scriptsize
 agriocnemis zerafica \textbf{EOT} 
 global distribution: the species is known from north-west uganda and 
 sudan, through niger to mauritania and liberia: a larger sahelian range, i.e.,~
 in more arid zone than other african agriocnemis. record from angola unlikely. 
 northeastern africa distribution: the species was listed by tsuda for sudan. [$\cdots$]. 
 \textbf{EOP} very small, about 20mm. 
 orange tail. advised agriocnemis sp. id by kd dijkstra: [$\cdots$] 
 \textbf{EOP} same creature as previously posted as unknown, 
 very small, about 20mm, over water, top view. advised probably agriocnemis, 
 "whisp" damselfly. \textbf{EOP} [$\cdots$] \textbf{EOP} 
 justification: this is a widespread species with no known major widespread threats 
 that is unlikely to be declining fast enough to qualify for listing in a threatened 
 category. it is therefore assessed as least concern. \textbf{EOP} the species has been 
 recorded from northwest uganda and sudan, through niger to mauritania and [$\cdots$] \textbf{EOP} the main 
 threats to the species are habitat loss due to agriculture, urban development and drainage, 
 as well as water pollution. 
 }
\end{tabular}  
}
\vspace{-0.5em}
\caption{Summary (top) and input paragraphs (bottom) from the Animal
  development dataset (EOP/T is a special token indicating the end of paragraph/title).}\label{tab:task}
\end{table*}

Although content structure has been largely unexplored within neural
text generation, it has been been recognized as useful for
summarization.  \citet{barzilay2004catching} build a model of the
content structure of source documents and target summaries and use it
to extract salient facts from the source.
\citet{sauper2009automatically} cluster texts by target topic and use
a global optimisation algorithm to select the best combination of
facts from each cluster.  Although these models have shown good
results in terms of content selection, they cannot generate target
summaries.  Our model is also related to the hierarchical decoding
approaches of \citet{lijiwei2015} and \citet{tanjiwei2017}.  However,
the former approach is auto-encoding the same inputs (our model
carries out content selection for the summarization task), while the
latter generates independent sentences. They also both rely on
recurrent neural models, while we use convolutional neural networks.
To our knowledge this is the first hierarchical decoder proposed for a
non-recurrent architecture.

To evaluate our model, we introduce \textsc{WikiCatSum}, a
dataset\footnote{Our dataset and code are available at
  \url{https://github.com/lauhaide/WikiCatSum}.}  derived from
\citet{liu2018generating} which consists of Wikipedia abstracts and
source documents and is representative of three domains, namely
Companies, Films, and Animals. In addition to differences in
vocabulary and range of topics, these domains differ in terms of the
linguistic characteristics of the target summaries.  We compare single
sequence decoders and structured decoders using ROUGE and a suite of
new metrics we propose in order to quantify the \emph{content
  adequacy} of the generated summaries.  
We also show that structured decoding improves content coverage based
on human judgments.

\section{The Summarization Task}
\label{sec:summarisation-task}

The Wikipedia lead section introduces the entity (e.g.,~\emph{Country}
or \emph{Brazil}) the article is about, highlighting important facts
associated with it.  \citet{liu2018generating} further assume that
this lead section is a summary of multiple documents related to the
entity.  Based on this premise, they propose the multi-document summarization
task of generating the  lead section from the set of
documents cited in Wikipedia articles or returned by Google (using
article titles as queries). And create WikiSum, a large-scale
multi-document summarization dataset with hundreds of thousands of
instances.

\citeauthor{liu2018generating} (\citeyear{liu2018generating}) focus on
summarization from very long sequences.  Their model first selects a
subset of salient passages by ranking all paragraphs from the set of
input documents (based on their {TF-IDF} similarity with the title of
the article).  The~$L$ best ranked paragraphs (up to 7.5k~tokens) are
concatenated into a flat sequence and a decoder-only architecture
\cite{vaswani2017attention} is used to generate the summary.

We explicitly model the topic structure of summaries, under the
assumption that documents cover different topics about a given entity,
while the summary covers the most salient ones and organizes them into
a coherent multi-sentence text.  We further assume that different lead
summaries are appropriate for different entities (e.g.~Animals
vs. Films) and thus concentrate on specific domains.  We associate
Wikipedia articles with ``domains'' by querying the DBPedia
knowledge-base.  A training instance in our setting is a
(domain-specific) paragraph cluster (\emph{multi-document} input) and
the Wikipedia lead section (target \emph{summary}).  We derive sentence 
topic templates from summaries for Animals, Films, and Companies and 
exploit these to guide the summariser. 
However, there is nothing inherent in our model that restricts its 
application to different domains.

\section{Generation with Content Guidance}
\label{sec:gener-with-cont}
Our model takes as input a set of ranked paragraphs ${\cal{P}}= \{ p_1
\cdots p_{|\cal{P}|} \}$ which we concatenate to form a flat input
sequence ${\cal{X}}= ( x_1 \cdots x_{|\cal{X}|} )$ where $x_i$ is the
$i$-th token.  The output of the model is a multi-sentence summary
${\cal{S}} = ( s_1, \cdots, s_{|\cal{S}|} )$ where $s_t$ denotes the
$t$-th sentence.

\begin{figure}
 \hspace{-1em}\includegraphics[scale=0.6]{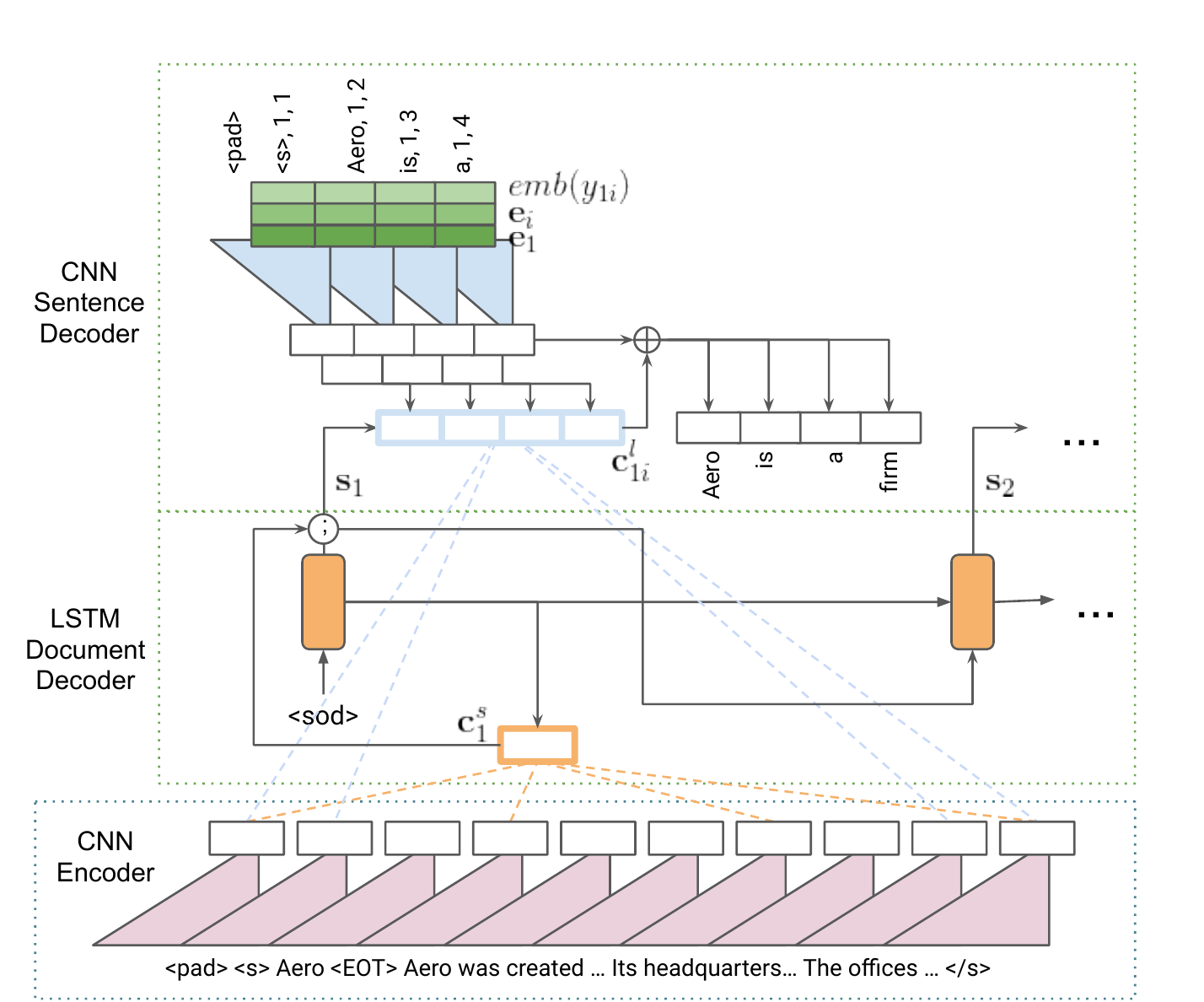}
\vspace{-1.8em}
\caption{Sequence encoder and structured decoder.}\label{fig:sdec-arch}
\end{figure}

We adopt an encoder-decoder architecture which makes use of
convolutional neural networks (CNNs;
\citealt{gehring2017convs2s}). CNNs permit parallel training
\cite{gehring2017convs2s} and have shown good performance in
abstractive summarization tasks (e.g., \citealt{narayan2018don}).
Figure~\ref{fig:sdec-arch} illustrates the architecture of our model.
We use the convolutional encoder of \citet{gehring2017convs2s} to
obtain a sequence of states $( \mathbf{z}_1, \cdots,
\mathbf{z}_{|\cal{X}|} )$ given an input sequence of tokens $( x_1,
\cdots, x_{|\cal{X}|} )$. A \emph{hierarchical} convolutional decoder
generates the target sentences (based on the encoder outputs).
Specifically, a \emph{document-level} decoder first generates sentence
vectors (LSTM Document Decoder in Figure~\ref{fig:sdec-arch}),
representing the content specification for each sentence that the
model plans to decode.  A \emph{sentence-level} decoder (CNN Sentence
Decoder in Figure~\ref{fig:sdec-arch}) is then applied to
generate an actual sentence token-by-token. In the following we
describe the two decoders in more detail and how they are combined to
generate summaries.

\subsection{Document-level Decoder}
The document-level decoder
builds a sequence of sentence representations ($\mathbf{s}_1, \cdots,
\mathbf{s}_{|\cal{S}|}$).  For example, $\mathbf{s}_1$ in
Figure~\ref{fig:sdec-arch} is the vector representation for the
sentence \textit{Aero is a firm}.  This layer uses an
LSTM with attention.  At each time step $t$, the LSTM will construct
an output state $ \mathbf{s}_t$, representing the content of the
$t$-th sentence that the model plans to generate:
\begin{gather}\label{eq:docdecoder}
 \mathbf{h}_t =  \text{LSTM}(\mathbf{h}_{t-1},\mathbf{s}_{t-1} )\\
  \mathbf{s}_t =  \text{tanh}(\mathbf{W}_s[\mathbf{h}_t;\mathbf{c}^s_{t}])
\end{gather}
where $\mathbf{h}_t$ is the LSTM  hidden state of  step $t$
and $\mathbf{c}^s_t$ is the  context vector
computed by attending to the input.
The initial hidden state $\mathbf{h}_{0}$ is initialized with
the averaged sum of the encoder output states.

We use a soft attention mechanism \cite{luong2015effective} 
to compute the  context vector $\mathbf{c}^s_{t}$:
\begin{gather}
   \alpha^s_{tj} =  \frac{\text{exp}(\mathbf{h}_t \, \bigcdot \, \mathbf{z}_j)}{\sum _{j^{\,\prime}} \text{exp}(\mathbf{h}_t \, \bigcdot \, \mathbf{z}_{j^{\,\prime}})}\\
   \mathbf{c}_t^s = \sum_{j=1}^{|\cal{X}|} \alpha^s_{tj} \, \mathbf{z}_j
\end{gather}
where $\alpha^s_{jt}$ is the attention weight for the document-level
decoder attending to input token $x_j$ at time step $t$.

\subsection{Sentence-level Decoder}
Each sentence $s_t=( y_{t1}, \dots, y_{t|s_t|} )$ in target
summary~$\cal{S}$ is generated by a sentence-level decoder.  The
convolutional architecture proposed in \citeauthor{gehring2017convs2s}
(\citeyear{gehring2017convs2s}) combines word embeddings with
positional embeddings.  That is, the word representation
$\mathbf{w}_{ti}$ of each target word $y_{ti}$ is combined with vector
$\mathbf{e}_i$ indicating where this word is in the sentence,
$\mathbf{w}_{ti} = emb(y_{ti}) + \mathbf{e}_i$.  We extend this
representation by adding a \emph{sentence positional embedding}.  For
each ~$s_t$ the decoder incorporates the representation of its
position~$t$.  This explicitly informs the decoder which sentence in
the target document to decode for. Thus, we redefine word
representations as
$\mathbf{w}_{ti} = emb(y_{ti}) + \mathbf{e}_i+ \mathbf{e}_t $.

\subsection{Hierarchical Convolutional Decoder}
In contrast to recurrent networks where initial conditioning
information is used to initialize the hidden state, in the
convolutional decoder this information is introduced via an attention
mechanism. In this paper we extend the multi-step attention
\cite{gehring2017convs2s} with sentence vectors~$\mathbf{s}_t$
generated by the document-level decoder.

The output vectors for each layer~$l$ in the convolutional decoder,
when generating tokens for the~\mbox{$t$-th} sentence
are\footnote{Padding and masking are used to keep the auto-regressive
  property in decoding.}:
\begin{gather}
\{\mathbf{o}^{l}_{t1},\cdots,\mathbf{o}^{l}_{tn}\} = \text{conv}(\{\mathbf{o'}^{l-1}_{t1},\cdots,\mathbf{o'}^{l-1}_{tn})\\
\mathbf{o'}^{l}_{ti} = \mathbf{o}^{l}_{ti}+\mathbf{s}_t+\mathbf{c}{^l_{ti}}
\end{gather}
where $\mathbf{o'}^{l}_{ti}$ is obtained by adding the corresponding
sentence state $\mathbf{s}_t$ produced by the document-level decoder
(Equation~(2)) and sentence-level context vector $\mathbf{c}_{ti}^l$.
$\mathbf{c}_{ti}^l$ is calculated by combining~$\mathbf{o}^{l}_{ti}$
and~$\mathbf{s}_t$ with the previous target
embedding~$\mathbf{g}_{ti}$:
\begin{gather}
\mathbf{d}^l_{ti} = W^l_d (\mathbf{o}_{ti}^{l} + \mathbf{s}_t) + \mathbf{g}_{ti} \\
a^l_{tij} = \frac{\text{exp}( \mathbf{d}^l_{ti} \bigcdot \mathbf{z}_j )} { \sum _{j^{\,\prime}} \text{exp}(\mathbf{d}^l_{ti} \bigcdot \mathbf{z}_{j^{\,\prime}} )} \\
\mathbf{c}{^l_{ti}} = \sum_{j=1}^{|\cal{X}|} a^l_{tij} (\mathbf{z}_j + \mathbf{e}_j)
\end{gather}

The prediction of word $y_{ti}$ is conditioned on the 
output vectors of the top convolutional layer, as 
$P(y_{ti}|y_{t\{1:i-1\}})=\softmax(W_y(\mathbf{o}^{L}_{ti}+\mathbf{c}{^L_{ti}}))$.
The model is trained to optimize negative log likelihood ${\mathcal L}_{NLL}$.

\subsection{Topic Guidance}

To further render the document-level decoder topic-aware, we annotate
the sentences of ground-truth summaries with topic templates and force
the model to predict these.  To discover topic templates from summaries, we train a Latent Dirichlet
Allocation model (LDA; \citet{Blei:2003:LDA}), treating sentences as
documents, to obtain sentence-level topic distributions.  Since the
number of topics discussed in the summary is larger than the number of
topics discussed in a single sentence, we use a symmetric Dirichlet
prior (i.e.,~we have no a-priori knowledge of the topics) with the
concentration parameter set to favour sparsity in order to encourage
the assignment of few topics to sentences.  We use the learnt topic
model consisting of ${\cal{K}} = \{ k_1, \cdots, k_{|\cal{K}|} \}$
topics to annotate summary sentences with a topic vector.  For each
sentence, we assign a topic label from $\cal{K}$ corresponding to its
most likely topic.  Table~\ref{tab:topics} shows topics discovered by
LDA and the annotated target sentences for the three domains we
consider.

\begin{table}[t]
\begin{footnotesize}
\centerline{
\begin{tabular}{@{}p{7.8cm}@{}} \thickhline
\multicolumn{1}{c}{Company}\\ \thickhline
\#12: operation, start, begin, facility, company, expand \\
\#29: service, provide, airline, member, operate, flight \\
\#31: product, brand, sell, launch, company, include \\
\#38: base, company, office, locate, development, headquarters \\
 \textit{Epos Now's UK headquarters are located in Norwich, England and their US headquarters are in Orlando, Florida.} [\textbf{\#38}]\\
\thickhline
\multicolumn{1}{c}{Film}\\\thickhline
\#10: base, film, name, novel, story, screenplay \\
\#14: win, film, music, award, nominate, compose \\
\#18: film, receive, review, office, box, critic \\
\#19: star, film, role, play, lead, support \\	
 \textit{The film is based on the novel Intruder in the dust by
   William Faulkner.} [\textbf{\#10}]\\\thickhline
 \multicolumn{1}{c}{Animal} \\ \thickhline
\#0: length, cm, reach, grow, centimetre, size, species \\
\#1: forewing, hindwing, spot, line, grey, costa \\
\#17: population, species, threaten, list, number, loss, endanger \\
\#24: forest, habitat, consist, area, lowland, moist, montane \\ 
  \textit{It might be in population decline due to habitat loss.}
  [\textbf{\#17}]\\ \thickhline
\end{tabular}
}
\end{footnotesize}
\vspace{-0.5em}
\caption{Topics discovered for different domains and examples of sentence annotations.}\label{tab:topics}
\end{table}

\begin{table}[t]
\centering
{\footnotesize
 \begin{tabular}{@{\extracolsep{\fill}}l@{\hspace{5pt}}c@{\hspace{5pt}}
 c@{\hspace{5pt}}c@{\hspace{5pt}}c@{\hspace{5pt}}c@{\hspace{5pt}}c@{\hspace{5pt}}c
 @{\hspace{5pt}}c@{\hspace{5pt}}c@{\hspace{5pt}}c}\thickhline
 Category & {InstNb} & R1 & R2 & RL & TopicNb\\
 \thickhline
 Company & 62,545 &  .551 & .217  & .438 & 40 \\ 
 Film    & 59,973 &  .559 & .243  & .456 & 20 \\ 
 Animal  & 60,816 &  .541 & .208  & .455 & 30 \\ \thickhline 
                               
\end{tabular}
}
\vspace{-0.5em}
\caption{Number of instances (InstNb), ROUGE~1-2 recall (R1 and R2) 
  of source texts against target summaries and number of topics (TopicNb).
}\label{tab:wikicatsum}
\end{table}

We train the document-level decoder to predict the topic $k_t$ of 
sentence $s_t$ as an auxiliary task,
$P(k_{t}|s_{1:t-1})=\softmax(W_k(\mathbf{s}_t))$, and optimize the
summation of the ${\mathcal L}_{NLL}$ loss and the negative log
likelihood
of~$P(k_{t}|s_{1:t-1})$. 

\section{Experimental setup}

\paragraph{Data}

Our \textsc{WikiCatSum} data set includes the first 800 tokens from
the input sequence of paragraphs \cite{liu2018generating} and the
Wikipedia lead sections.  We included pairs with more than 5 source
documents and with more than 23 tokens in the lead section (see Appendix~\ref{sec:supplemental} 
for details). 
Each dataset was split into train (90\%), validation (5\%) and test set
(5\%). Table~\ref{tab:wikicatsum} shows dataset statistics.

We compute recall ROUGE scores of the input documents against the 
summaries to asses the amount of overlap and as a reference for the
interpretation of the scores achieved by the models.  Across domains
content overlap (R1) is~\textasciitilde50 points. However, R2 is much
lower indicating that there is abstraction, paraphrasing, and content
selection in the summaries with respect to the input.  We rank input
paragraphs with a \emph{weighted} TF-IDF similarity metric which takes
paragraph length into account \cite{singhal2017pivoted}.

The column TopicNb in Table~\ref{tab:wikicatsum} shows the number of
topics in the topic models selected for each domain and
Table~\ref{tab:topics} shows some of the topics (see
Appendix~\ref{sec:supplemental} for training and selection details). 
The optimal number of topics differs for
each domain.  In addition to general topics which are discussed across
domain instances (e.g.,~topic \#0 in Animal), there are also more
specialized ones, e.g., relating to a type of company (see topic \#29
in Company) or species (see topic \#1 in Animal).

\paragraph{Model Comparison}
We compared against two baselines: the Transformer
sequence-to-sequence model (\textsc{TF-S2S}) of
\citeauthor{liu2018generating} (\citeyear{liu2018generating}) and the
Convolutional sequence-to-sequence model (\textsc{CV-S2S}) of
\citeauthor{gehring2017convs2s} (\citeyear{gehring2017convs2s}).
\textsc{CV-S2D} is our variant with a single sequence encoder and a
structured decoder; and +\emph{T} is the variant with topic label
prediction.  TF-S2S has 6 layers, the hidden size is set to 256 and
the feed-forward hidden size was 1,024 for all layers.
All convolutional models use the same encoder and decoder convolutional
blocks.  The encoder block uses 4 layers, 256 hidden dimensions and
stride 3; the decoder uses the same configuration but 3 layers. All
embedding sizes are set to 256.  \textsc{CV-S2D} models are trained by
first computing all sentence hidden states $\mathbf{s}_t $ and then
decoding all sentences of the summary in parallel. See Appendix~\ref{sec:supplemental}
for models training details.

At test time, we use beam size of 5 for all models. The structured
decoder explores at each sentence step 5 different hypotheses.
Generation stops when the sentence decoder emits the
\textbf{E}nd-\textbf{O}f-\textbf{D}ocument (EOD) token.  The model
trained to predict topic labels, will predict the
\textbf{E}nd-\textbf{O}f-\textbf{T}opic label.  This prediction is
used as a hard constraint by the document-level decoder, setting the
probability of the EOD token to~1. We also use trigram blocking
\cite{paulus2017deep} to control for sentence repetition and discard consecutive
sentence steps when these overlap on more than 80\% of the tokens. 

\begin{table}[t]
\centering
\begin{footnotesize}
\begin{tabular}{@{\extracolsep{\fill}}l@{\hspace{2pt}}c@{\hspace{2pt}}c@{\hspace{2pt}}c|@{\hspace{2pt}}c
@{\hspace{2pt}}c@{\hspace{2pt}}c|@{\hspace{2pt}}c@{\hspace{2pt}}c@{\hspace{2pt}}c}
\thickhline
   Model &\multicolumn{3}{c}{Company} & \multicolumn{3}{c}{Film} & \multicolumn{3}{c}{Animal} \\
   & R1 & R2 & RL & R1 & R2 & RL & R1 & R2 & RL \\

\thickhline
\textsc{TF-S2S} & .260 & .095 & .204 & .365 & .188 & .310 & \textbf{.440} & \textbf{.288} & \textbf{.400}\\
\textsc{CV-S2S} & .245 & .094 & .199 & .346 & .198 & .307 & .422 & .284 & .385 \\
\textsc{CV-S2D} & \textbf{.276} & .105 & .213 & .377 & .208 & .320 & .423 & .273 & .371 \\
\textsc{CV-S2D}\emph{+T} & .275 & \textbf{.106} & \textbf{.214} & \textbf{.380} & \textbf{.212} & \textbf{.323} & .427 & .279 & .379 \\
\thickhline
&  & A & C &  & A & C &  & A & C \\
\thickhline
\textsc{CV-S2S} & & .046 & .307 & &  .097 & .430 & &  \textbf{.229} & \textbf{.515} \\
\textsc{CV-S2D} & & .051 & .314 & &  .098 & .429 & &  .219 & .499 \\
\textsc{CV-S2D}\emph{+T} & & \textbf{.051} & \textbf{.316} & &  \textbf{.101} & \textbf{.433} &  & .223 & .506 \\
\thickhline
\end{tabular} 
\end{footnotesize}
\vspace{-0.5em}
\caption{ROUGE F-scores (upper part) and additional content metrics (bottom part). }
\label{tab:rouges-extra}
\end{table} 

\section{Results}
\label{sec:results}

\paragraph{Automatic Evaluation}
Our first evaluation is based on the standard ROUGE
metric \cite{lin:2004:ACLsummarization}.  We also
make use of two additional automatic metrics.  They are based on
unigram counts of content words and aim at quantifying how much the
generated text and the reference overlap with respect to the input
\cite{sari2016}. We expect multi-document summaries to cover details
(e.g., names and dates) from the input but also abstract and rephrase
its content.  \emph{Abstract}~(A) computes unigram f-measure between
the reference and generated text excluding tokens from the input.  
Higher values indicate the model's abstraction capabilities.
\emph{Copy} (C) computes unigram f-measure between the reference and
generated text only on their intersection with the input. Higher
values indicate better coverage of input details.

Table~\ref{tab:rouges-extra} summarizes our results on the test set.
In all datasets the structured decoder brings a large improvement in
ROUGE-1 (R1), with the variant using topic labels (\emph{+T}) bringing
gains of~+2 points on average.  With respect to ROUGE-2 and -L (R2 and
RL), the \textsc{CV-S2D}\emph{+T} variant obtains highest scores on
Company and Film, while on Animal it is close below to the baselines.
Table~\ref{tab:rouges-extra} also presents results with our additional
metrics which show that \textsc{CV-S2D} models have a higher overlap
with the gold summaries on content words which do not appear in the
input (A). All models have similar scores with respect to content
words in the input and reference (C).

\paragraph{Human Evaluation}
We complemented the automatic evaluation with two human-based
studies carried out on Amazon Mechanical Turk (AMT) over 
45~randomly selected examples from the test set (15 from each
domain). We compared the 
\textsc{TS-S2S}, \textsc{CV-S2S} and \textsc{CV-S2D}\emph{+T} models.

The first study focused on assessing the extent to which generated
summaries retain salient information from the input set of
paragraphs. We followed a question-answering (QA) scheme as proposed
in \citet{clarke2010discourse}.  Under this scheme, a set of questions
are created based on the gold summary; participants are then asked to
answer these questions by reading system summaries alone without
access to the input. The more questions a system can answer, the
better it is at summarizing the input paragraphs as a whole (see
Appendix~\ref{sec:supplemental} for example questions).
Correct answers are given a score of 1, partially correct answers score 0.5, 
and zero otherwise. The final score is the average of all question scores.
We created between two and four factoid questions for each summary;
a total of 40 questions for each domain. We collected 3	judgements per
system-question	pair. Table~\ref{tab:humaneval} shows the QA scores.
Summaries by the \textsc{CV-S2D}\emph{+T} model are able to answer
more questions, even for the Animals domain where the \textsc{TS-S2S} model
obtained higher ROUGE scores.

\begin{table}
\centering
\begin{footnotesize}
\begin{tabular}{@{\extracolsep{\fill}}l@{\hspace{2pt}}l@{\hspace{2pt}}c|@{\hspace{2pt}}l@{\hspace{2pt}}c|
@{\hspace{2pt}}l@{\hspace{2pt}}c}
\thickhline
Model &\multicolumn{2}{c}{Company} & \multicolumn{2}{c}{Film} & \multicolumn{2}{c}{Animal} \\
& QA & Rank & QA & Rank & QA & Rank\\
\thickhline
  \textsc{TF-S2S} & 5 & 1.87 & 6 & 2.27 & 9 & 1.87 \\
\textsc{CV-S2S} & 5 & 2.27 & 6.67 &  1.76 & 8.33 & 2.04 \\
\textsc{CV-S2D}\emph{+T} & 7 & 1.87 & 7 & 1.98 & 9.33 & 2.09 \\
\thickhline
 \end{tabular} 
 \end{footnotesize}
\vspace{-0.5em}
\caption{QA-based evaluation and system ranking.}\label{tab:humaneval}
\end{table} 

The second study assessed the overall content and linguistic quality
of the summaries. We asked judges to rank (lower rank is better) system outputs according to
\emph{Content} (does the summary appropriately captures the content of 
the reference?), \emph{Fluency} (is the summary fluent and grammatical?),
\emph{Succinctness} (does the summary avoid repetition?).  We
collected 3 judgments for each of the 45 examples.  Participants were
presented with the gold summary and the output of the three systems in
random order.  
Over all domains, the ranking of the \textsc{CV-S2D}\emph{+T} model is better
than the two single-sequence models \textsc{TS-S2S} and \textsc{ConvS2S}.

\section{Conclusions}

We introduced a novel structured decoder module for multi-document
summarization.  Our decoder is aware of which topics to mention in a
sentence as well as of its position in the 
summary. Comparison of our model against competitive single-sequence decoders
shows that structured decoding yields summaries
with better content coverage.

\section*{Acknowledgments}

We thank the ACL reviewers for their constructive feedback.
We gratefully acknowledge the financial support of the European
Research Council (award number 681760).

\bibliographystyle{acl_natbib}
\bibliography{protosum}

\appendix

\section{Appendix}
\label{sec:supplemental}

\subsection{Data}

WikiSum consist of Wikipedia articles each of which are associated
with a set of reference documents.\footnote{We take the processed
  Wikipedia articles from
  \url{https://github.com/tensorflow/tensor2tensor/tree/master/tensor2tensor/data_generators/wikisum}
  released on April 25th 2018.}  We associate Wikipedia articles
(i.e.,~entities) with a set of categories by querying the DBPedia
knowledge-base.\footnote{Entities of Wikipedia articles are associated
  with categories using the latest DBPedia release
  \url{http://wiki.dbpedia.org/downloads-2016-10} to obtain the
  instance types
  (\url{http://mappings.dbpedia.org/server/ontology/classes/}).}  The
WikiSum dataset originally provides a set of URLs corresponding to the
source reference documents; we crawled online for these references
using the tools provided in \citet{liu2018generating}.\footnote{The
  crawl took place in July 2018 and was supported by Google Cloud Platform.}

\begin{table}
\centering
\begin{footnotesize}
 \begin{tabular}{@{\extracolsep{\fill}}l@{\hspace{5pt}}c@{\hspace{5pt}}c} \thickhline
 Category & SentNb & SentLen \\ \thickhline
 Company & 5.09$\pm$3.73 & 24.40$\pm$13.47  \\ 
 Film    & 4.17$\pm$2.71 & 23.54$\pm$11.91  \\ 
 Animal  & 4.71$\pm$3.53 & 19.68$\pm$18.69  \\
\thickhline
\end{tabular}
\end{footnotesize}
\caption{Average number of sentences in target summaries (SentNb)
and sentence length (SentLen) in terms of word counts.
}\label{tab:wikicatsum:full}
\end{table} 
  
We used the Stanford CoreNLP \cite{manning-EtAl:2014:P14-5} to
tokenize the lead section into sentences. We observed that the Animal
data set contains overall shorter sentences but also sentences
consisting of long enumerations which is reflected in the higher
variance in sentence length (see SentLen in
Table~\ref{tab:wikicatsum:full}).  An example (lead) summary and
related paragraphs in shown in Table~\ref{fig:task:full}. The upper part
shows the target summary and the bottom the input set of
paragraphs. \textbf{EOP} tokens separate the different paragraphs,
\textbf{EOT} indicates the title of the Wikipedia article.

To discover sentence topic templates in summaries, we used the Gensim
framework \cite{rehurek_lrec} and learned LDA models on summaries of the 
train splits.  We performed grid search on the number of topics $[10,
\cdots, 90]$ every ten steps, and used the context-vector-based topic coherence metric
(cf. \cite{cv:topic:coherence}) as guidance to manually inspect the output topic
sets and select the most appropriate ones.  For competing topic sets,
we trained the models and selected the topic set which led to higher
ROUGE scores on the development set.

We used the following hyperparameters to train topic models with
Gensim \cite{rehurek_lrec}.  We set the $\alpha=0.001$ and
$\eta=\text{'auto'}$; and used the following training configuration:
{\ttfamily random\_state=100, eval\_every=5,
    chunksize=10000, iterations=500, passes=50}.
We train on the preprocessed version of the summaries with lemmas 
of content words (stop words were removed).

\subsection{Model Training Details}
In all convolutional models we used dropout \cite{SrivastavaHKSS14} in
both encoder and sentence-level decoder with a rate of 0.2. For 
the normalisation and initialisation of the convolutional architectures, we follow \cite{gehring2017convs2s}.
Similarly, to train the convolutional models we follow the optimisation setup in \cite{gehring2017convs2s}.

For the transformer-based baseline we applied dropout (with probability 
of 0.1) before all linear layers and label smoothing \cite{SzegedyVISW16} 
with smoothing factor 0.1. The optimizer was Adam \cite{kingma2015adam} with 
learning rate of 2, $ \beta_1 = 0.9$, and $ \beta_2 = 0.998$; we also applied 
learning rate warm-up over the first 8,000 steps, and decay as in \cite{vaswani2017attention}. 

We select the best models based on ROUGE scores on the development set.

As for the data, we discarded examples where the lead 
contained sentences longer than 200 tokens (often been long 
enumerations of items).  For the training of all models
we only retained those data examples fitting the maximum target
length of the structured decoder, 15 sentences with maximum 
length of 40 tokens (sentences longer than this where split).
We used a source and target vocabulary of 50K words for all datasets.

On decoding we normalise log-likelihood of the candidate hypotheses $y$
by their length, $|y|^\alpha$ with $\alpha=1$ \cite{wuetal2016},
except for the structured decoder on the Animals dataset where
we use $\alpha=0.9$. For the transformer model we use $\alpha=0.6$.

\begin{table*}[h]
{\small
\centerline{
\begin{tabular}{p{14cm}}\thickhline
 agriocnemis zerafica is a species of damselfly in the family coenagrionidae.   
 it is native to africa, where it is widespread across the central and western 
 nations of the continent. it is known by the common name sahel wisp.   
 this species occurs in swamps and pools in dry regions.   
 there are no major threats but it may be affected by pollution and habitat loss 
 to agriculture and development. \\\thickhline
 agriocnemis zerafica \textbf{EOT} specimen count 1 record last modified 21 apr 2016 nmnh -entomology dept. 
 taxonomy animalia arthropoda insecta odonata coenagrionidae 
 collector eldon h. newcomb preparation envelope prep count 1 sex male stage 
 adult see more items in specimen inventory entomology place area 5.12km. ne. 
 dakar, near kamberene; 1:30-4:30 p.m., senegal collection date 21 may 
 1944 barcode 00342577 usnm number usnment342577 published name agriocnemis zerafica 
 le roi \textbf{EOP} global distribution: the species is known from north-west uganda and 
 sudan, through niger to mauritania and liberia: a larger sahelian range, i.e.,~
 in more arid zone than other african agriocnemis. record from angola unlikely. 
 northeastern africa distribution: the species was listed by tsuda for sudan. this 
 record needs confirmation. may also occur in kenya as well. \textbf{EOP} very small, about 20mm. 
 orange tail. advised agriocnemis sp. id by kd dijkstra: hard to see details, but i 
 believe this is not a. exilis \textbf{EOP} same creature as previously posted as unknown, 
 very small, about 20mm, over water, top view. advised probably agriocnemis, 
 "whisp" damselfly. \textbf{EOP} thank you for taking the time to provide feedback on the 
 iucn red list of threatened species website, we are grateful for your input. \textbf{EOP} 
 justification: this is a widespread species with no known major widespread threats 
 that is unlikely to be declining fast enough to qualify for listing in a threatened 
 category. it is therefore assessed as least concern. \textbf{EOP} the species has been 
 recorded from northwest uganda and sudan, through niger to mauritania and liberia: 
 a larger sahelian range, i.e.,~in more arid zone than other african \textbf{EOP} the main 
 threats to the species are habitat loss due to agriculture, urban development and drainage, 
 as well as water pollution. 
 \textbf{EOP} no conservation measures known but information on
 taxonomy, population ecology, habitat status and population trends
 would be valuable. \\ \thickhline
\end{tabular}  
}
}
\caption{Summary (top) and input paragraphs (bottom) from the Animal
  development dataset.}\label{fig:task:full}
\end{table*}

\begin{table*}[h]
\centering
\begin{small}
 \begin{tabular}{@{~}p{1.5cm}p{12.5cm}@{~}}
  \thickhline
\multicolumn{2}{c}{Film} \\  \thickhline
Gold & Mary Queen of Scots is a 2013 Swiss period drama directed by Thomas Imbach.  
It is his first film in English and French language starring the bilingual french actress Camille Rutherford.  
The film portrays the inner life of Mary, the Queen of Scotland. The film is based on austrian novelist 
Stefan Zweig's 1935 biography, Mary Stuart, a long-term bestseller in Germany and France but out of 
print in the UK and the us for decades until 2010. The film was first screened at the 2013 
International Film Festival Locarno and was later shown at the 2013 Toronto International Film Festival.\\
\multicolumn{2}{c}{} \\
\multirow{3}{*}{QA}
&What does the film portrays? \hspace{1ex}[the inner life of Mary , the Queen of Scotland]\\
&At which festival was the film first screened? \hspace{1ex}[2013 International Film Festival Locarno]\\
&Who is the author of the novel the film is based on? \hspace{1ex}[Stefan Zweig]\\

\multicolumn{2}{c}{} \\
\textsc{TF-S2S} & Mary Queen of Scots is a 2013 British biographical film based on the life 
of Mary Queen Mary Mary Queen of Scots. It was directed by Ian Hart and stars Vanessa Redgrave 
as the title role. It was released in the United Kingdom on 18 april 2013.\\[2.5em]
\textsc{CV-S2S} & Mary Queen of Scots is a 2013 German drama film directed by Thomas UNK. It 
was screened in the contemporary world cinema section at the 2013 Toronto International Film Festival.\\[1.5em]
\makecell{\textsc{CV-S2D}\emph{+T}} & Mary Queen of Scots ( german : das UNK der UNK ) is a 2013 
German drama film directed by Thomas UNK. The film is based on the life of Mary Ellen of Scots. 
The film was released in the united states on January 17 , 2013.\\

\thickhline 

 \end{tabular} 
\end{small}

\caption{Example of Gold summary, question set and system outputs for the QA evaluation study.}\label{tab:qa-output}
\end{table*}

\subsection{Evaluation and System Outputs}

In the automatic evaluation we used {\ttfamily pyrouge}\footnote{\url{pypi.python.org/pypi/pyrouge}}
and ROUGE-1.5.5.pl with stemming (parameters= ``{\ttfamily -c 95 -r 1000 -n 2 -m}'').

Table~\ref{tab:qa-output} shows an example of gold summary and corresponding question
set from the question-answering study in Section~\ref{sec:results}.
Table~\ref{tab:sys-output-additional} shows examples of system output
on the development set. Specifically, we show summaries generated by 
\textsc{ConvS2S} and \textsc{ConvS2D}\emph{+Topic}, and also include
the reference Gold standard.

\begin{table*}[h]
\centering
\begin{small}
 \begin{tabular}{@{~}p{1.5cm}p{12.5cm}@{~}}
  \thickhline
\multicolumn{2}{c}{Company} \\  \thickhline
Gold & Seagull Book, formerly called Seagull Book \& Tape, is an American retail chain bookstore focusing 
on products for members of the Church of Jesus Christ of latter-day Saints (lds church), with over two 
dozen stores in Utah, Idaho, Arizona, and nevada. It was the second largest lds bookstore until being acquired 
in 2006 by market-leader deseret book, and since then Seagull has continued to operate as a discount chain, 
distinct from deseret book branded retail stores.\\[4.5em]
\textsc{CV-S2S} & Seagull Book \& Tape, Inc. is a book publishing company based in american fork, Utah, United States. 
It was founded in 1987 by jonathan UNK.\\[1.4em]
\makecell{\textsc{CV-S2D}\emph{+T}} & Seagull Book \& Tape, Inc. is an American book retailer with 26 stores 
throughout Utah, Idaho and California. The company is based in Boise, Idaho. The company is based in Boise, idaho, 
with its sister company Seagull Book \& Tape.\\
\thickhline  
\multicolumn{2}{c}{} \\
\multicolumn{2}{c}{} \\
\multicolumn{2}{c}{} \\\thickhline
\multicolumn{2}{c}{Film} \\ \thickhline
Gold & To Write Love on Her Arms (also known as Day One; formerly Renee) is a 2012 american biographical drama 
film written and directed by Nathan Frankowski, starring Kat Dennings, Chad Michael Murray, Rupert Friend, 
Juliana Harkavy, Corbin Bleu and Mark Saul. The film is based on the life of troubled teenager Renee Yohe 
and the founding of To Write Love on Her Arms by Jamie Tworkowski, after he and others helped Yohe to overcome 
her challenges enough to be able to enter rehab. The film premiered on march 11, 2012 at the Omaha Film Festival, 
and was eventually released direct-to-dvd on March 3, 2015.\\[6.7em]
\textsc{CV-S2S} & To UNK Love on Her Arms is a 2015 American biographical drama film directed by Renee UNK 
and written by Renee UNK. The film is based on the true story of a girl whose journey is threatened by her arms. \\[2.6em]
\makecell{\textsc{CV-S2D}\emph{+T}} & To Write Love on Her Arms is a 2015 American biographical drama film 
directed by Renee UNK. The film is based on the true story of Renee UNK. The film was released in the United 
States on March 3, 2015. The film is based on the book of the same name by Renee UNK.\\
\thickhline  
\multicolumn{2}{c}{} \\
\multicolumn{2}{c}{} \\
\multicolumn{2}{c}{} \\ \thickhline
\multicolumn{2}{c}{Animal} \\ \thickhline
Gold & Compacta Capitalis is a moth in the Crambidae family. It was described by Grote in 1881. It is found 
in North America, where it has been recorded from Maryland to Florida, West to Texas and possibly Colorado, 
North to Illinois. The wingspan is about 35 mm. The forewings are forewing are white with a reddish-brown 
shading at the base and along the inner margin and two black discal spots, as well as an irregular subterminal line.
There is a dark apical blotch on both wings. Adults are on wing from May to August. \\[5.7em]
\textsc{CV-S2S} & Compacta UNK is a moth in the Crambidae family. It was described by Barnes and McDunnough in 1918. 
It is found in North America, where it has been recorded from Alabama, Florida, Georgia, Illinois, Indiana, Kentucky, 
Maine, Maryland, Massachusetts, Minnesota, New Brunswick, New Hampshire, New Jersey, New york, North Carolina, Ohio, 
Oklahoma, Ontario, Pennsylvania, Quebec, South Carolina, Tennessee, Texas and Virginia.\\[4.6em]
\makecell{\textsc{CV-S2D}\emph{+T}} & Compacta UNK is a moth in the Crambidae family. It was described by Grote in 1878. 
It is found in North America, where it has been recorded from Florida. It is also found in Mexico. The wingspan 
is about 20 mm. Adults have been recorded on wing from April to September.\\
\thickhline  
 \end{tabular} 
\end{small}

\caption{Examples of system output on the development set.}\label{tab:sys-output-additional}
\end{table*}

\end{document}